\documentclass[preprint,5p,twocolumn]{elsarticle}

\usepackage{amssymb}
\usepackage{amsmath}

\usepackage{graphicx}
\graphicspath{{images/}}

\usepackage{subcaption}

\usepackage[]{algorithm2e}

\usepackage{hyperref}

\journal{ArXiv.org}

\begin{document}


\newcommand{\myfigure}[3] {
    \begin{figure}
        \centering
        \includegraphics[width=1.0\columnwidth]{#1}
        \caption{#2}
        \label{#3}
    \end{figure}
}

\newcommand{\vect}[1]{\boldsymbol{#1}}

\begin{frontmatter}


\title{Super-klust: Another Way of Piecewise Linear Classification}

\author[1,3]{Rahman Salim Zengin
\corref{cor1}
\fnref{fn1}}
\ead{rszengin@itu.edu.tr}

\author[2,3]{Volkan Sezer
\fnref{fn2}}
\ead{sezerv@itu.edu.tr}

\cortext[cor1]{Corresponding author}

\fntext[fn1]{\href{https://orcid.org/0000-0002-3104-4677}{ORCID: https://orcid.org/0000-0002-3104-4677}}

\fntext[fn2]{\href{https://orcid.org/0000-0001-9658-2153}{ORCID: https://orcid.org/0000-0001-9658-2153}}

\address[1]{Department of Mechatronics Engineering, Istanbul Technical University, Istanbul, Turkey}
\address[2]{Department of Control and Automation Engineering, Istanbul Technical University, Istanbul, Turkey}
\address[3]{Autonomous Mobility Group, Electrical and Electronics Engineering Faculty, Istanbul Technical University, Istanbul, Turkey}

\begin{abstract}
    With our previous study, the Super-k algorithm, we have introduced a novel
    way of piecewise-linear classification. While working on the Super-k
    algorithm, we have found that there is a similar, and simpler way to explain
    for obtaining a piecewise-linear classifier based on Voronoi tessellations.
    Replacing the multidimensional voxelization and expectation-maximization
    stages of the algorithm with a distance-based clustering algorithm,
    preferably k-means, works as well as the prior approach. Since we are
    replacing the voxelization with the clustering, we have found it meaningful
    to name the modified algorithm, with respect to Super-k, as
    \textbf{Supervised k Clusters} or in short \textbf{Super-klust}. Similar to
    the Super-k algorithm, the Super-klust algorithm covers data with a labeled
    Voronoi tessellation, and uses resulting tessellation for classification.
    According to the experimental results, the Super-klust algorithm has similar
    performance characteristics with the Super-k algorithm.
\end{abstract}


\end{frontmatter}

\section{Introduction}
\label{sec:introduction}

Piecewise-linear (PWL) classifiers gives us an alternate way of handling
classification problems. With our previous study, the Super-k
algorithm\cite{zengin_super-k_2021}, we have introduced a novel approach to PWL
classification problem. During further development efforts of Super-k we have
found that, using a distance based clustering method, the preferred one is
k-means\cite{jain_data_2010}, we can replace multidimensional voxelization and
expectation-maximization (EM) stages without losing functionality of the
algorithm. When Super-k analyzed with the perspective of clustering, the
multidimensional voxelization works as initialization, and the EM works as
clustering.

Using a well known clustering algorithm for the first stages of our novel PWL
classifier increases clarity of the algorithm and takes the burden of managing a
part of the complexity from our shoulders. With the introduction of distance
based clustering into the algorithm, it is more meaningful to name algorithm
with respect to clustering. So, we are introducing a modified version of the
Super-k algorithm and naming it as \textbf{Supervised k Clusters} or in short
\textbf{Super-klust}.

In this paper, we will explain the differences of the Super-klust algorithm and
give experimental results similar, and comparable to the Super-k algorithm. In
the following section (Section \ref{sec:super-klust}) the Super-klust algorithm
is explained. In Section \ref{sec:experimental}, graphical results on synthetic
datasets are given, and the Super-klust algorithm is compared with Super-k,
SVM-linear, SVM-RBF, SVM-polynomial, and KNN using the experimental results of
the Super-k paper\cite{zengin_super-k_2021}. The results are also discussed. In
Section \ref{sec:conclusion}, the paper is summed up and the final thoughts are
given.

\section{The Super-klust Algorithm}
\label{sec:super-klust}

The Super-klust algorithm starts with clustering of separate classes of the
data. The clustering algorithm of choice is k-means\cite{jain_data_2010}.
Although it is possible to use different $k$ values for different classes, in
order to keep the implementation simple, $k$ is a shared parameter between
clustering of different classes.

Clustering of different classes of the data into $k$ clusters creates separate
Voronoi tessellations for every class. With the k-means algorithm, $k$ cluster
means becomes the generator points that represent that class as a Voronoi
tessellation. One advantage of the k-means algorithm, although it depends on the
implementation, it gives us the ability to control the number of generator
points finely via the $k$ parameter.

\begin{figure}[h]
    \centering
    \includegraphics[width=\linewidth]{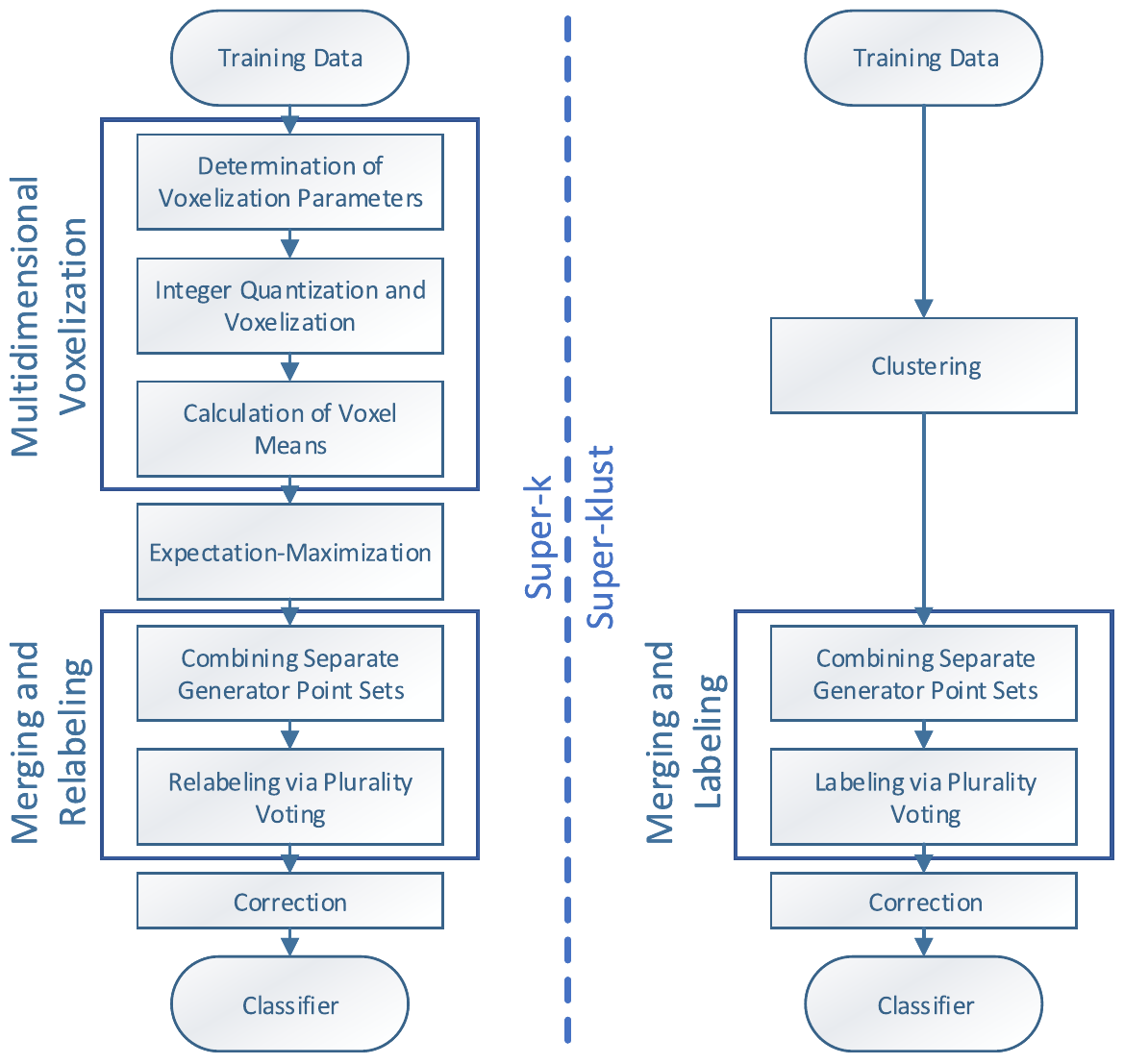}
    \caption{The Super-k and Super-klust algorithms}
    \label{fig:superk_vs_superklust}
\end{figure}

After obtaining the sets of generator points separately for the classes, the
remaining part of the algorithm is same as the Super-k algorithm. The details
can be found in our prior paper\cite{zengin_super-k_2021}. The process steps of
the Super-k and Super-klust algorithms are shown in (Figure
\ref{fig:superk_vs_superklust}).

\section{Experimental Results}
\label{sec:experimental}

The platform used for the experimental tests and comparisons is as follows: The
CPU used for the experimentation is Intel(R) Core(TM) i7-7700 running at 3.60GHz
frequency; the system has 32GB of DDR4 RAM running at 2133 MHz.

Unoptimized reference implementation of the Super-klust algorithm is written in
Python\cite{van_rossum_python_1995} using NumPy\cite{van_der_walt_numpy_2011} as
the linear algebra library and Scikit-Learn\cite{pedregosa_scikit-learn_2011}
for the k-means clustering. For the generation of the figures,
Matplotlib\cite{hunter_matplotlib_2007} is used.
Scikit-Learn\cite{pedregosa_scikit-learn_2011} is also used for both synthetic
data generation and dataset retrieval. In addition, the SVM variants and KNN are
used from the same library\cite{pedregosa_scikit-learn_2011}.

The reference implementation of the Super-klust algorithm and the source code to
reproduce the results of this paper are
shared\cite{rahman_salim_zengin_ituamgsuper-klust_2021}.

\subsection{Tests with synthetic datasets}
\label{sub:synthetic_tests}

The tests on synthetic datasets are done to visualize the behavior of the
Super-klust algorithm. The datasets are the same datasets, used in the Super-k
paper. Due to the finer controllability of the $k$ parameter, simpler
classifiers can be obtained via the Super-klust algorithm.

\begin{figure*}
    \centering
    \begin{subfigure}[b]{0.32\textwidth}
        \centering
        \includegraphics[width=\textwidth]{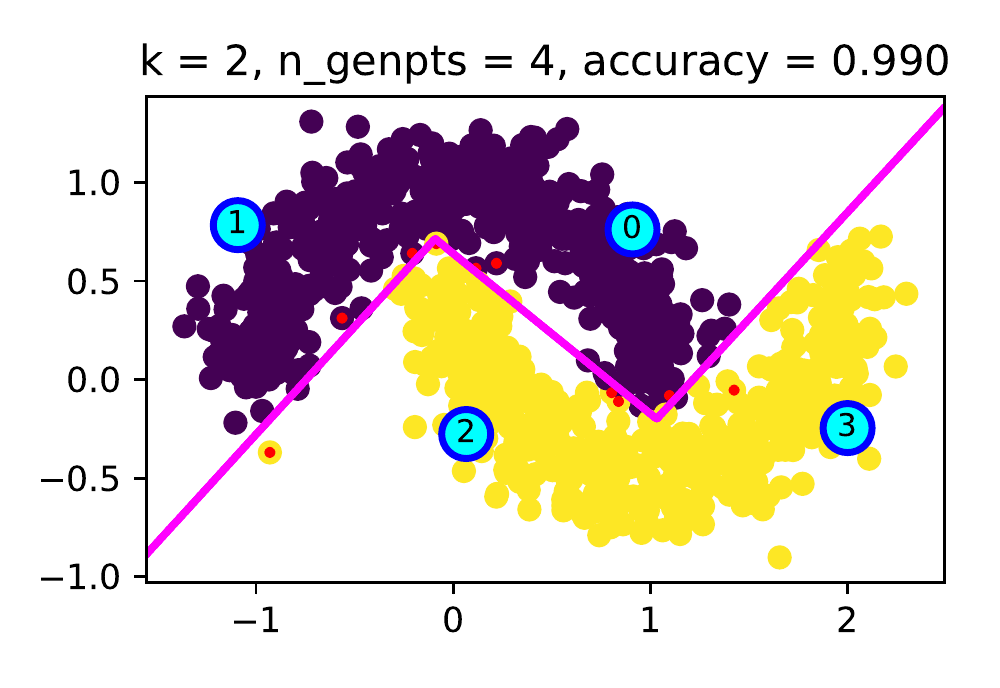}
        \caption{$k=3$}
        \label{fig:test_moons:1}
    \end{subfigure}
    \begin{subfigure}[b]{0.32\textwidth}
        \centering
        \includegraphics[width=\textwidth]{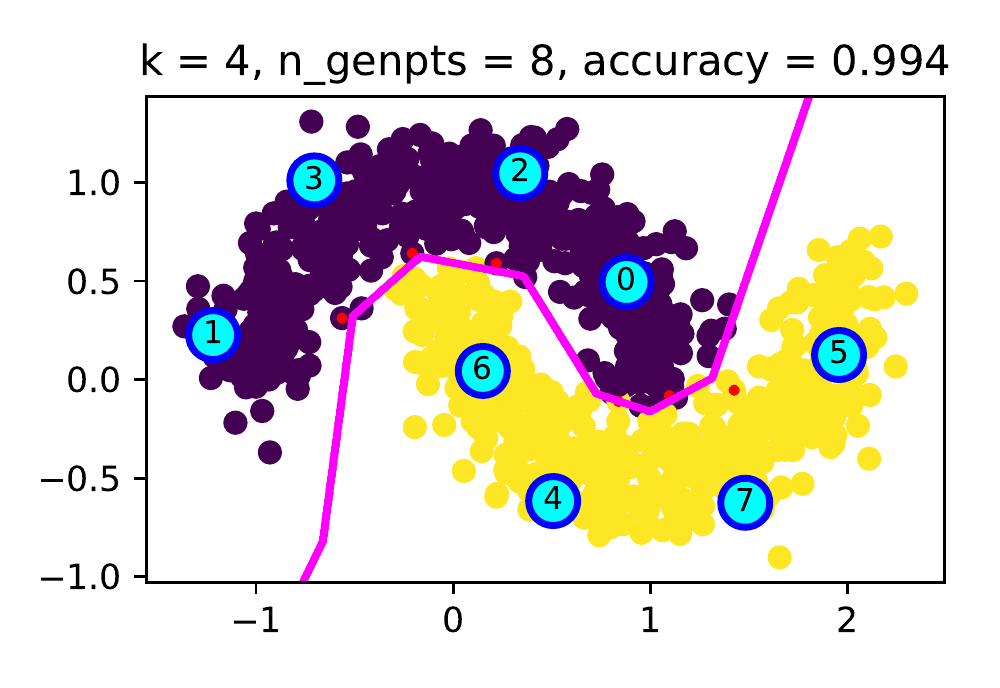}
        \caption{$k=10$}
        \label{fig:test_moons:2}
    \end{subfigure}
    \begin{subfigure}[b]{0.32\textwidth}
        \centering
        \includegraphics[width=\textwidth]{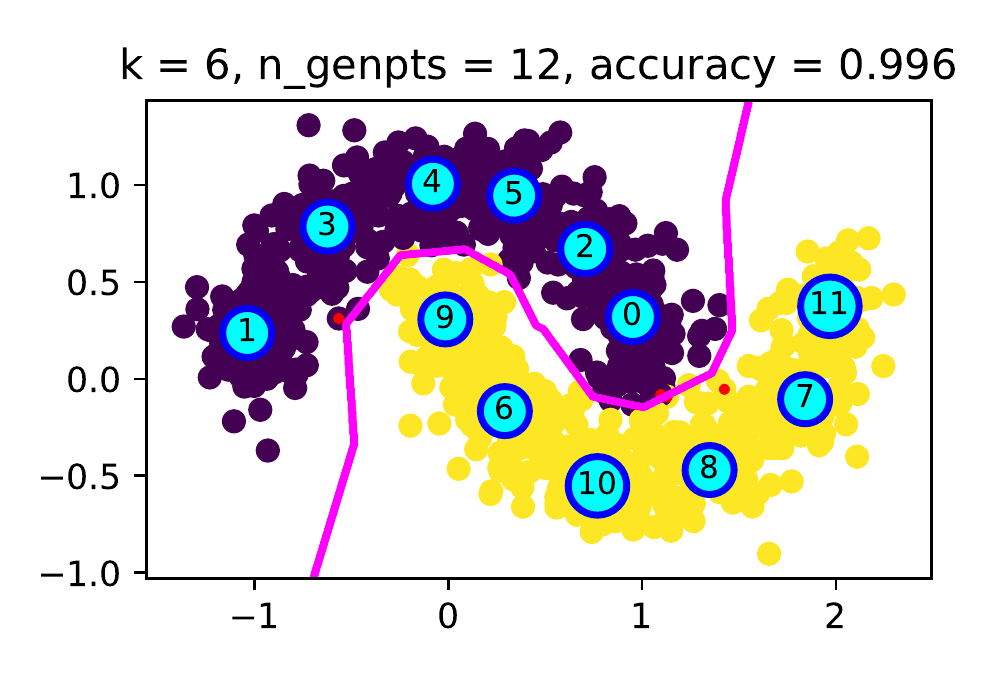}
        \caption{$k=17$}
        \label{fig:test_moons:3}
    \end{subfigure}
    \caption{Test results on synthetic moons dataset}
    \label{fig:test_moons}
\end{figure*}

As it is showed in (Figure \ref{fig:test_moons:1}), with $k = 2$, a minimalistic
classifier can be obtained. And with the Super-klust algorithm, $k$ directly
reflects the number of generator points per class. This can be clearly observed
in the (Figure \ref{fig:test_moons}).

\begin{figure*}
    \centering
    \begin{subfigure}[b]{0.32\textwidth}
        \centering
        \includegraphics[width=\textwidth]{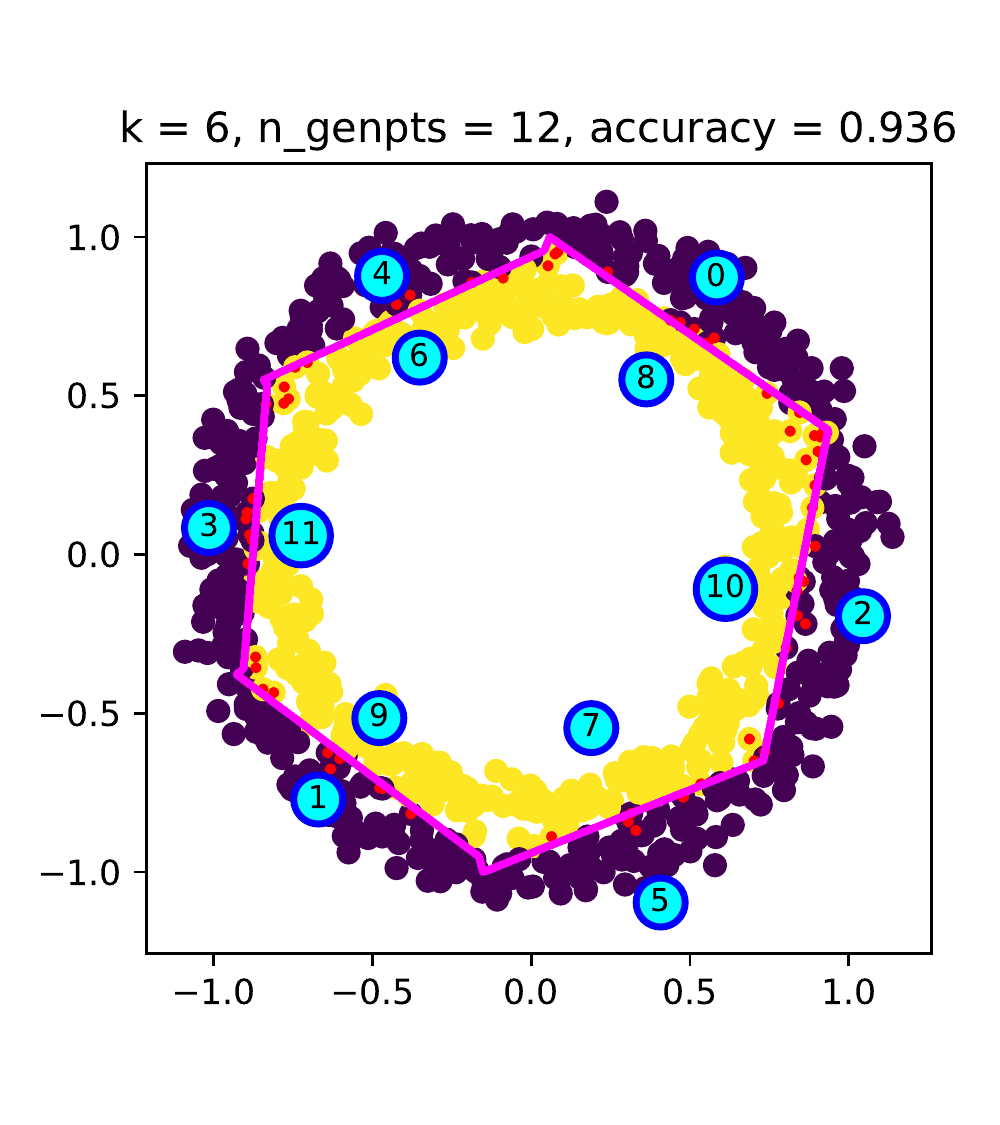}
        \caption{$k=10$}
        \label{fig:test_circles:1}
    \end{subfigure}
    \begin{subfigure}[b]{0.32\textwidth}
        \centering
        \includegraphics[width=\textwidth]{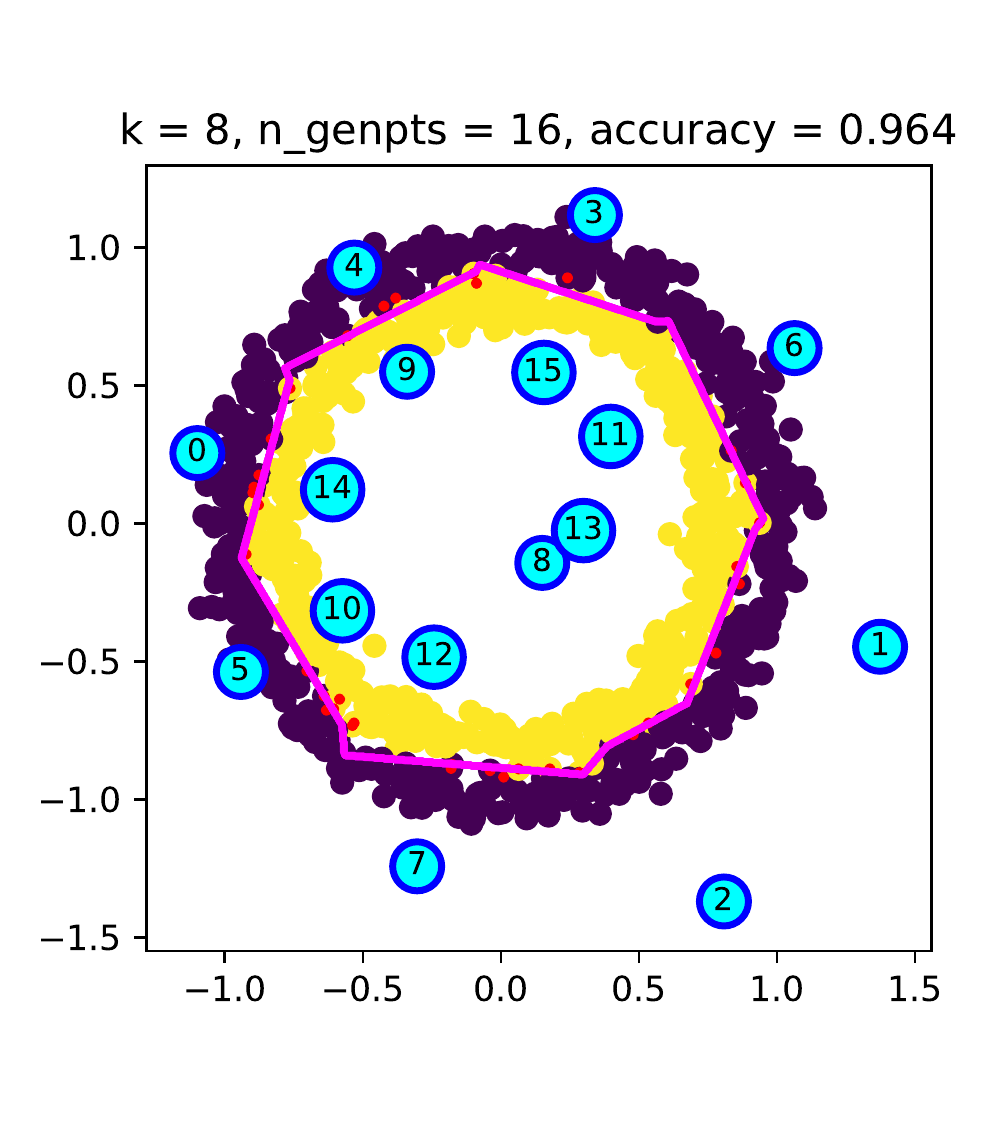}
        \caption{$k=20$}
        \label{fig:test_circles:2}
    \end{subfigure}
    \begin{subfigure}[b]{0.32\textwidth}
        \centering
        \includegraphics[width=\textwidth]{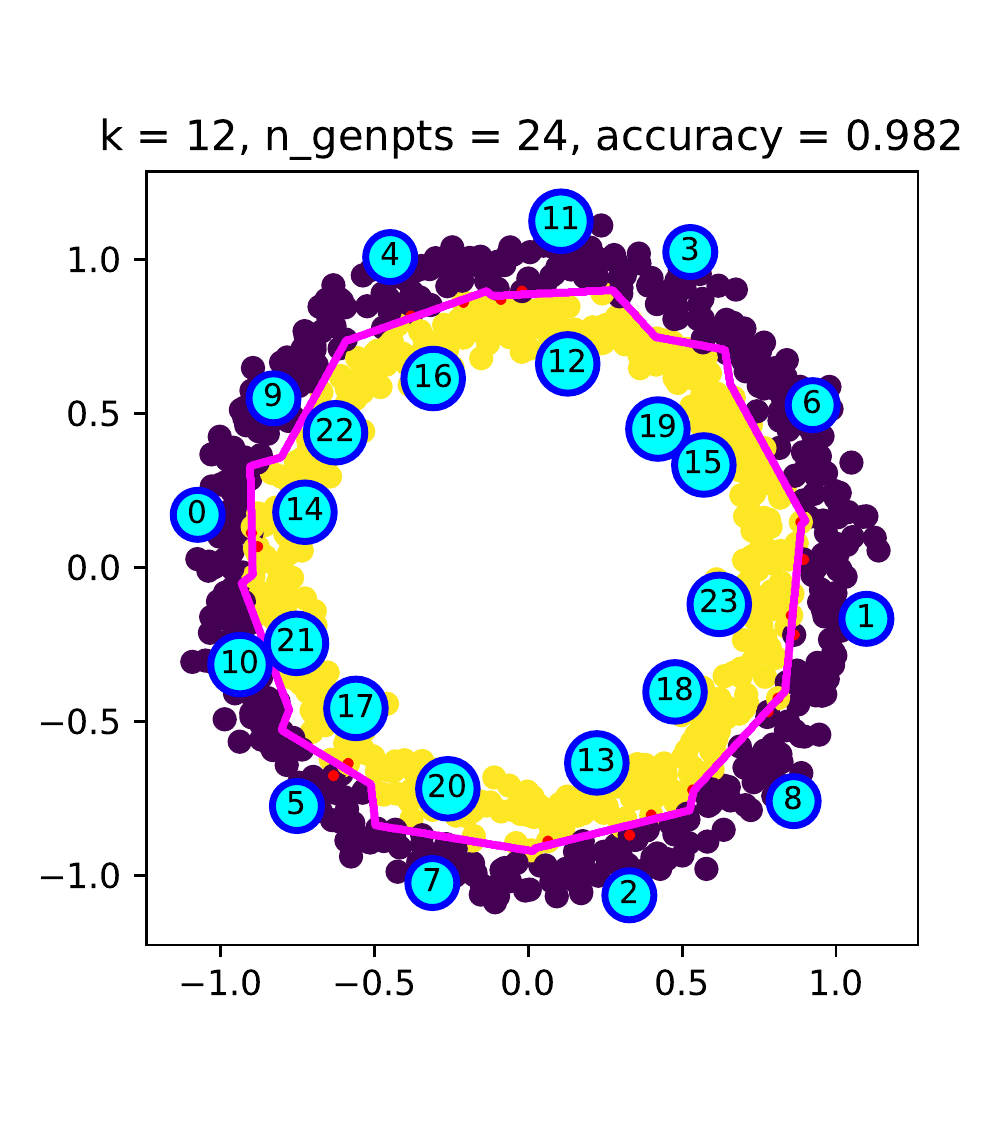}
        \caption{$k=30$}
        \label{fig:test_circles:3}
    \end{subfigure}
    \caption{Test results on synthetic circles dataset}
    \label{fig:test_circles}
\end{figure*}

Due to the random initialization of the k-means algorithm, the generator points
are occurring in random order. Although it is not so meaningful for the
classification, the orientations of the classifiers in (Figure
\ref{fig:test_circles}) looks random when they are compared to the figures of
the Super-k paper. And the random orientations shows importance of the
correction stage of the process.

\begin{figure*}
    \centering
    \begin{subfigure}[b]{0.32\textwidth}
        \centering
        \includegraphics[width=\textwidth]{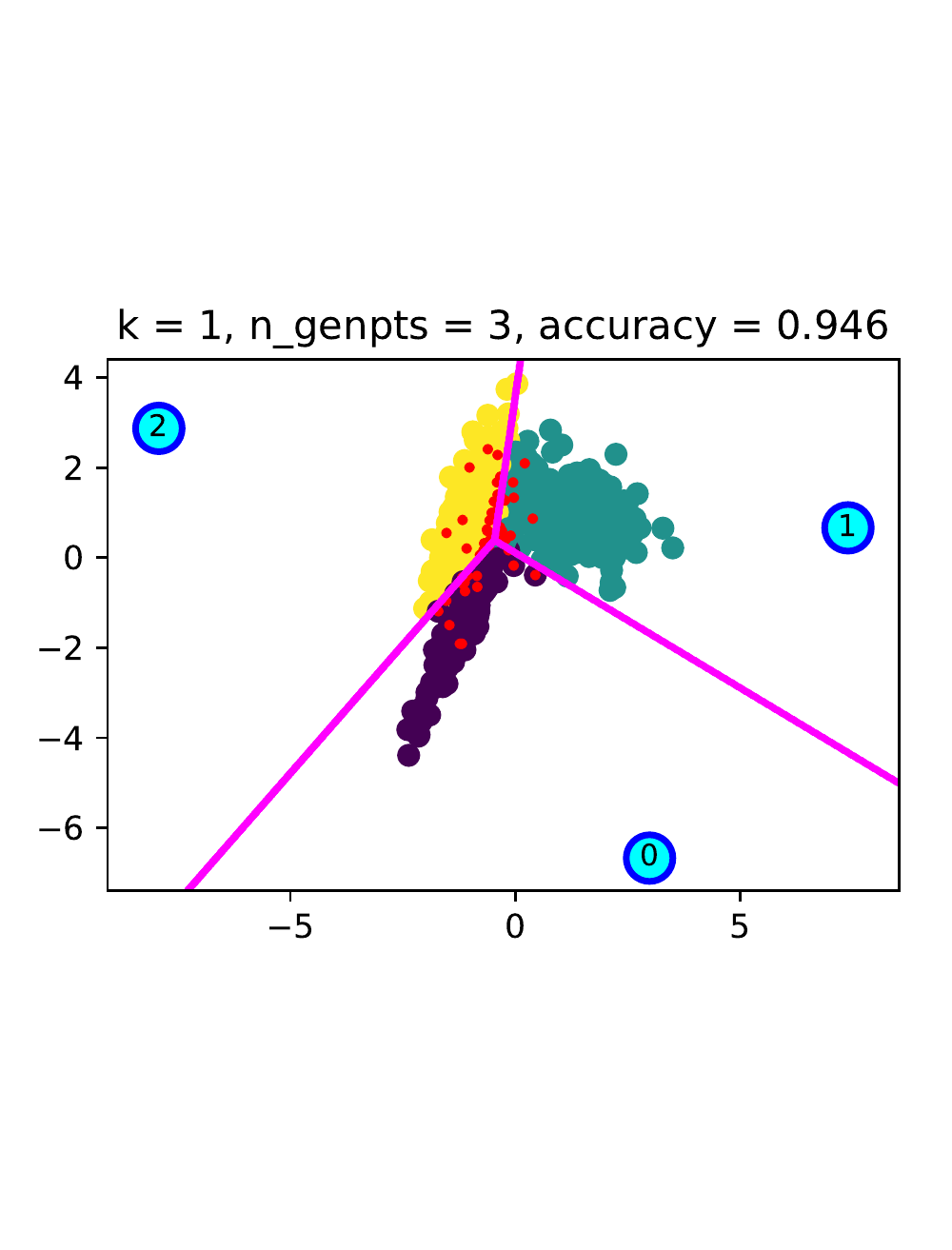}
        \caption{$k=2$}
        \label{fig:test_random:1}
    \end{subfigure}
    \begin{subfigure}[b]{0.32\textwidth}
        \centering
        \includegraphics[width=\textwidth]{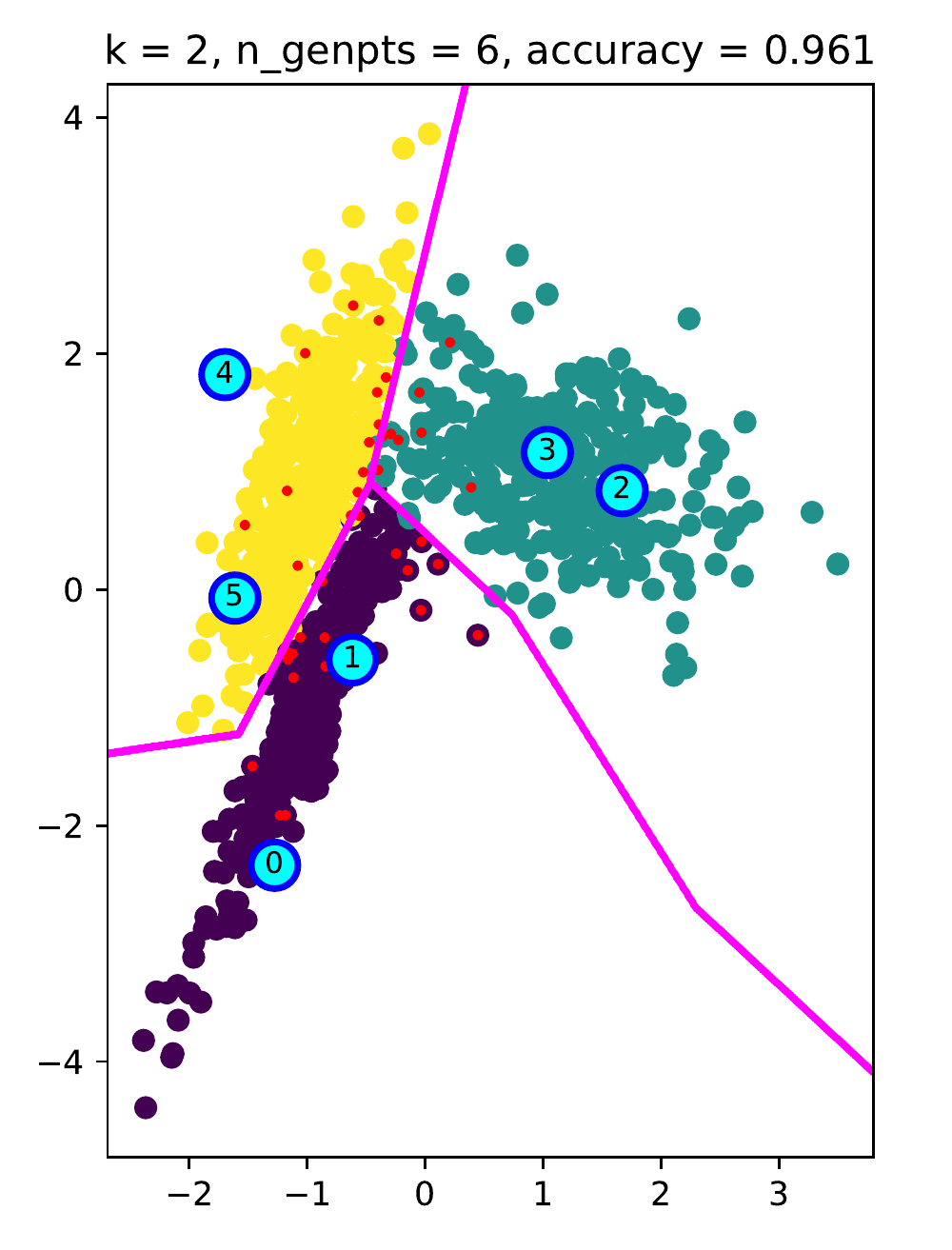}
        \caption{$k=5$}
        \label{fig:test_random:2}
    \end{subfigure}
    \begin{subfigure}[b]{0.32\textwidth}
        \centering
        \includegraphics[width=\textwidth]{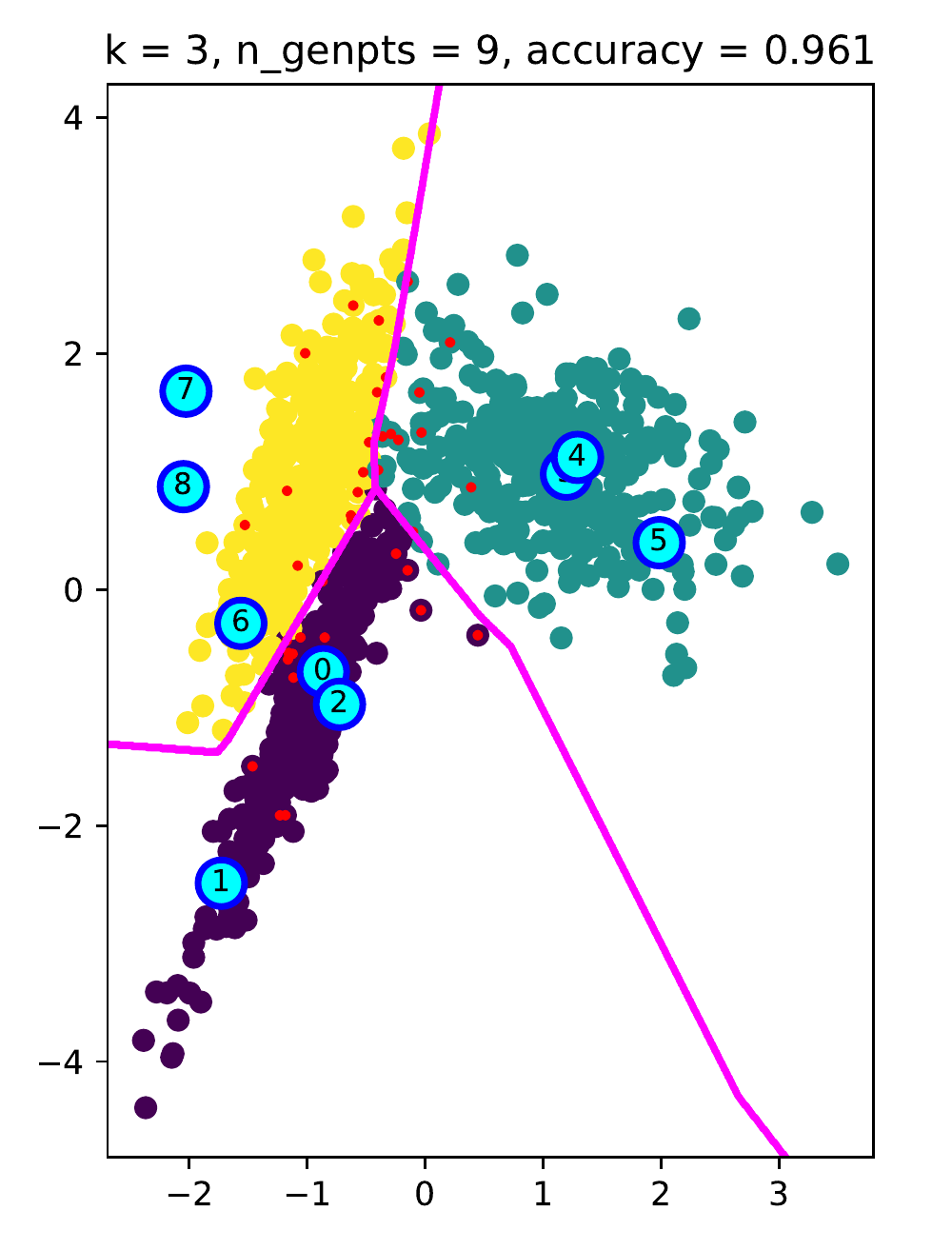}
        \caption{$k=8$}
        \label{fig:test_random:3}
    \end{subfigure}
    \caption{Tests results on randomly generated gaussians}
    \label{fig:test_random}
\end{figure*}

For the random gaussians dataset, increasing the k parameter does not have much
effect after $k = 2$, as it can be seen in (Figure \ref{fig:test_random}).

\subsection{Tests with real world datasets}
\label{sub:comparative_tests}

As it can be seen in the (Table \ref{tbl:experimental}), the results of the
Super-klust algorithm are similar to the results of the Super-k algorithm. The
differences are highly dependent on the parameters which are controlling the
algorithms. The Super-klust algorithm has comparable accuracy and training
performance with the others and shares the high performance inference
ability of the Super-k algorithm.

\begin{table*}
    \centering
    \caption{Experimental results with real world datasets}
    \begin{subtable}[t]{\textwidth}
        \centering
        \caption{Datasets}
        \begin{tabular}{|r|rrr|}
            \hline
                        & Sample Size & Features & Train/Test Sizes \\
            \hline
            optdigits   & 5620        & 64       & 3823/1797        \\
            USPS        & 9298        & 256      & 7291/2007        \\
            satimage    & 6430        & 36       & 5144/1286        \\
            letter      & 20000       & 16       & 16000/4000       \\
            isolet      & 7797        & 617      & 6240/1557        \\
            \hline
        \end{tabular}
        \label{tbl:experimental:datasets}        
    \end{subtable}
    ~\\
    \begin{subtable}[t]{\textwidth}
        \centering
        \caption{Test accuracies}
        \begin{tabular}{|r|rrrrr|}
            \hline
                       & optdigits & USPS  & satimage & letter & isolet \\
            \hline
            Super-klust & 0.968     & 0.938 & 0.909    & 0.950  & 0.938  \\
            Super-k     & 0.967     & 0.941 & 0.916    & 0.931  & 0.945  \\
            SVM Linear  & 0.961     & 0.926 & 0.877    & 0.850  & 0.962  \\
            SVM RBF     & 0.976     & 0.949 & 0.903    & 0.927  & 0.960  \\
            SVM Poly    & 0.976     & 0.954 & 0.866    & 0.947  & 0.965  \\
            KNN         & 0.979     & 0.949 & 0.911    & 0.955  & 0.923  \\
            \hline
        \end{tabular}
        \label{tbl:experimental:test_accuracies}        
    \end{subtable}
    ~\\
    \begin{subtable}[t]{\textwidth}
        \centering
        \caption{Training times in milliseconds, Mean{\tiny (StdDev)}}
        \begin{tabular}{|l|lllll|}
            \hline
                       & optdigits & USPS  & satimage & letter & isolet \\
            \hline
            Super-klust & 172.5{\small (7.0)} & 1287.4{\small (105.0)} & 259.7{\small (14.3)} & 2734.1{\small (272.7)} & 833.0{\small (147.9)}  \\
            Super-k     & 101.3{\small (0.2)} & 1922.8{\small (855.4)} & 177.5{\small (2.3)}  & 1402.7{\small (190.9)} & 1633.9{\small (289.4)} \\
            SVM Linear  & 101.4{\small (2.0)} & 1440.7{\small (78.9)}  & 393.3{\small (51.2)} & 3744.4{\small (155.5)} & 4917.7{\small (172.6)} \\
            SVM RBF     & 228.3{\small (0.6)} & 2718.6{\small (147.9)} & 378.0{\small (20.0)} & 3062.3{\small (240.5)} & 8636.5{\small (171.2)} \\
            SVM Poly    & 114.4{\small (0.4)} & 1901.9{\small (139.9)} & 423.9{\small (20.9)} & 2467.9{\small (174.0)} & 8203.0{\small (833.6)} \\
            KNN         & 13.9{\small (0.1)}  & 147.5{\small (6.5)}    & 18.1{\small (0.0)}   & 59.2{\small (3.5)}     & 291.8{\small (14.9)}   \\
            \hline
        \end{tabular}
        \label{tbl:experimental:training_times}
    \end{subtable}
    ~\\
    \begin{subtable}[t]{\textwidth}
        \centering
        \caption{Inference times in milliseconds, Mean{\tiny (StdDev)}}
        \begin{tabular}{|l|lllll|}
            \hline
                       & optdigits & USPS  & satimage & letter & isolet \\
            \hline
            Super-klust & 0.4{\small (0.0)}   & 3.7{\small (0.4)}      & 0.7{\small (0.0)}   & 19.5{\small (0.2)}     & 2.7{\small (0.0)}       \\
            Super-k     & 0.4{\small (0.0)}   & 3.2{\small (0.0)}      & 0.5{\small (0.0)}   & 6.8{\small (0.5)}      & 2.0{\small (0.1)}       \\
            SVM Linear  & 53.8{\small (0.1)}  & 543.4{\small (25.1)}   & 76.3{\small (0.1)}  & 1143.3{\small (30.5)}  & 3641.0{\small (187.3)}  \\
            SVM RBF     & 160.6{\small (0.6)} & 1207.4{\small (89.0)}  & 140.1{\small (7.8)} & 2424.3{\small (325.6)} & 5132.4{\small (222.6)}  \\
            SVM Poly    & 74.1{\small (0.1)}  & 823.7{\small (33.1)}   & 106.7{\small (5.5)} & 961.9{\small (41.9)}   & 4599.7{\small (151.8)}  \\
            KNN         & 472.5{\small (1.0)} & 4996.3{\small (173.8)} & 155.0{\small (8.9)} & 573.6{\small (29.4)}   & 10692.0{\small (122.7)} \\
            \hline
        \end{tabular}
        \label{tbl:experimental:test_times}
    \end{subtable}
    \label{tbl:experimental}
\end{table*}

\section{Conclusion}
\label{sec:conclusion}

The new foundational PWL classification algorithm, \textbf{Super-k}, is
simplified via introducing clustering as the method of obtaining separate
Voronoi tessellations for every class of the data. As a result,
\textbf{Supervised k Clusters}, or in short \textbf{Super-klust} becomes a new
option as a PWL classifier for our current and future research.

The contributions of the proposed algorithm are as follows:

\begin{itemize}
    \item A new and simpler PWL classification algorithm is introduced.
    \item The tremendous inference performance advantage of the Super-k
    algorithm is kept for the Super-klust algorithm.
\end{itemize}

According to the experimental results, the Super-klust algorithm has a
tremendous inference performance advantage over other algorithms, similar to
Super-k. The Super-klust algorithm shared all advantages and the opportunities
of the Super-k algorithm, with a simpler algorithm structure.

\section*{Acknowledgements}
\label{sec:acknowledgements}

This research was supported by the Turkish Scientific and Technological Research
Council (TUBITAK) under project no. 118E809.

We would like to thank the reviewers for their thoughtful comments and their
constructive remarks.

\newpage

\bibliographystyle{elsarticle-num} 
\bibliography{super-klust_paper.bib}

\end{document}